\documentclass[submission,copyright,creativecommons]{eptcs}
\providecommand{\event}{ICLP DC 2021}
\usepackage[utf8]{inputenc}
\usepackage{amsmath}
\usepackage{amssymb}
\usepackage{microtype}
\usepackage{url}
\usepackage{graphicx}
\usepackage{listings}

\def\titlerunning{Product Configuration in Answer Set Programming}
\def\authorrunning{S. Mishra}

\begin{document}

\title{Product Configuration in Answer Set Programming}

\author{Seemran Mishra\\
  University of Potsdam, Germany
}

\maketitle

\begin{abstract}
    \textbf{Abstract.} This is a preliminary work on configuration knowledge representation which serves as a foundation for building interactive configuration systems in Answer Set Programming (ASP).
    The major concepts of the product configuration problem are identified and discussed with a bike configuration example.
    A fact format is developed for expressing product knowledge that is domain-specific and can be mapped from other systems. 
    Finally, a domain-independent ASP encoding is provided that represents the concepts in the configuration problem.
\end{abstract}
 \section{Introduction}\label{sec:introduction}
Product configuration is one of the most successful commercial applications of artificial intelligence techniques \cite{soinie99a} \cite{falsch14a}.
It has been used from telephone switching systems \cite{stfrha98a}, to smart home configurations \cite{fefaateruraz17a}, in cement manufacturing plants \cite{orsben14a}, and the automotive industry \cite{tiheanso13a}. 
Being highly dynamic, this field evolved from mass production to mass customisation \cite{tisonisu03a}, as the customer needs became more individualistic.
The recent trend is interactive configuration \cite{fahakrscscta20a}, where the user is involved in every step of the configuration process.

ASP \cite{gellif88b} is a declarative modeling approach used for product configuration. 
In fact, product configuration was one of the first practical applications of ASP \cite{sonitisu01a}. 
Since then, there have been many works that use ASP to represent product configuration \cite{gekasc11c} \cite{mytirafe14a} \cite{fefaateruraz17a}.
But most of them are either domain-specific or tackle a subset of the configuration problem. 
Others express product knowledge in form of ASP rules.
This requires domain-experts to be proficient in ASP, which is not always the case.

This work serves as a foundation for interactive configuration systems in ASP.
The motivation here is the identification of the major concepts related to product configuration and the corresponding representation in ASP.
The development of a domain-independent configuration knowledge representation is another motivation.
This can be achieved by a clear separation between product knowledge and configuration knowledge.
The advantage here is that the configuration knowledge can be reused for any product to be configured and ensures better system maintenance. 
Stress is given for representing product knowledge as facts, to facilitate seamless mapping from other systems.
This enables domain-experts to provide product knowledge without being experts in ASP \cite{zhang14a}.

This paper is organized as follows: In Section 2, a background on product configuration is given.
The configuration problem is described in Section 3 using bike configuration as an example.
An ASP-based representation of the configuration problem and the corresponding solution is given in Section 4. 
Finally, the future work is expressed in Section 5.
 \section{Background}\label{sec:background}

One of the earliest applications of product configuration was R1/XCON, built on a rule-based representation language, where the solving knowledge was intertwined with domain-specific configuration knowledge \cite{mcdermott82b}.  
The trend quickly evolved to model-based knowledge representation where there is a clear separation of problem solving knowledge from domain-specific configuration knowledge.
As the industry shifted from mass production to mass customisation, another boundary between domain-specific configuration knowledge and user requirements was drawn \cite{mitfra89a}.

To create a general understanding of configuration, \cite{sotimasu98a} introduced concepts like \textit{component types}, \textit{properties} and constraints like \textit{partof} that are discussed in detail in the Section 3.
Also, the configuration is classified to connection-based, structure-based, resource-based, and function-based approaches.
Most concepts used in this work is adapted from \cite{sotimasu98a} and \cite{hofestrybawo14a}.

Representation and modeling of configuration knowledge have been split into three categories, namely constraint-based, graphical and logic-based knowledge representation \cite{hofestrybawo14a}.
Constraint programming has been one of the major used techniques for product configuration. 
Starting with \textit{static constraint satisfaction} where all the variables had to be assigned values, the switch was done to \textit{dynamic constraint satisfaction} \cite{mitfal90a}, where the configuration variables can be activated and deactivated during the search process.  
Owing to the lack of support for component-oriented modelling in the previous techniques, \textit{generative constraint satisfaction} (GCSP) \cite{stfrha98a} emerged.
GCSP was designed for product configuration and has widespread uses in the industry.

The main graphic-based representation systems used for constraint satisfaction are feature models and UML component diagrams \cite{mailharro98a}, \cite{fefrjastza03a}. 
These systems improve the accessibility of configuration knowledge to domain experts who are not acquainted with modeling, since these systems are more understandable and maintainable.
Predicate, as well as description logic, have been used for representing configuration problems, along with a mapping from UML diagrams \cite{fefrjastza03a}. 

ASP has been used widely in product configuration \cite{sonitisu01a} \cite{tiheanso13a}.
Linux Package Configuration in \cite{gekasc11c} is another interesting application of ASP.
An ASP representation of a configuration problem in the context of feature models is given in \cite{mytirafe14a}.
Mapping of object-oriented concepts to ASP directed at product configuration in UML was shown in \cite{faryscsh15a}. 
In \cite{fefaateruraz17a}, an introductory example of representing smart home configurations in ASP is provided.
 \section{Product Configuration Problem}\label{sec:current}

In this section, some of the major concepts used in product configuration are described in the context of bike configuration.
The building blocks of a configuration problem are \textit{components}. 
The components in the bike configuration problem consist of a bike, a frame, two wheels namely the front wheel and the rear wheel, a stand, and a basket as shown in Figure \ref{fig:bike}.
The domain of each component in the configuration problem is a set of \textit{component type}, shortened to \textit{type}.
Examples of types include w1, w2, and w3 which are in the domain of both the front wheel and the rear wheel. 

A component is characterized by a set of \textit{properties}.
The types also have \textit{properties}, whose values are predefined.
For example, the material of frame type f1 is aluminum.
The domain of component properties apart from type is given by the predefined property values of the corresponding types of components.
As an example in the bike configuration example, the material of the frame can be either carbon fiber or aluminum. 
To make the representation more uniform, type is considered as a property of the component. 

A \textit{configuration problem} can be defined as a set of components with properties, a set of domains for each component property, and a set of constraints listed below: 
\begin{itemize}
    \item \textbf{Property Assignment}: These constraints deal with the assignment of values to component properties from their respective domains.
    \begin{itemize}
        \item \textbf{A1}: Every component present in the configuration solution must be assigned a type.
        \item \textbf{A2}: A component property can only have one value. 
        \item \textbf{A3}: The values assigned to a component property must exactly correspond to the predefined property values of the type it is assigned.
        \item \textbf{A4}: All mandatory properties of a component should be assigned values if the component is in a configuration solution.
    \end{itemize}    
    \item \textbf{Partonomy}: The structure of a product can be represented as a partonomy where a whole component may have optional and mandatory parts. 
    For example, a bike is a whole component while a frame is a mandatory part and a basket is an optional part of the bike.
    \begin{itemize}
        \item \textbf{P1}: If a part component is present in an assignment, the whole component must be also in the configuration. 
        \item \textbf{P2}: If a component is present in an assignment, all the mandatory part components have to be in the assignment.    
    \end{itemize}
    \item \textbf{Requirements}: These constraints specify that including certain components (with property values) forces other components (with property values) to be present in the configuration.
    \begin{itemize}
        \item \textbf{R1}: A component requires another component to be part of the solution.
        \item \textbf{R2}: To satisfy the configuration, a component requires another component with a specific property value to be assigned to the configuration (or vice versa). 
        \item \textbf{R3}: To satisfy the configuration, a component with a certain property value requires another component with a specific property value to be assigned to the configuration. 
    \end{itemize} 
    \item \textbf{Incompatibility}: These constraints specify that certain combinations of components (with property values) are not allowed in the configuration solution.
    \begin{itemize}
        \item \textbf{I1}: Incompatible components cannot be together in a configured solution. 
        \item \textbf{I2}: A component can be incompatible with a certain presence value of another component. In this case, if the former component is present in the solution, then the later component can't have the respective property value. 
        \item \textbf{I3}: The property value of a component can be incompatible with the property value of another component. In this case, only one of the components and their property value can be part of the solution or both have to be excluded. 
    \end{itemize}
    \item \textbf{User Requirements}: The users can specify their requirements in terms of components or specific values of component properties.
    \begin{itemize}
        \item \textbf{U1}: Every component that the user requests, must be part of the configuration.
        \item \textbf{U2}: A user can require a component with a specific property value. In this case, the specified component with the respective property value must be present in the solution. 
        \item \textbf{U3}: Every component that the user requests not to be present, must not be part of the configuration.
        \item \textbf{U4}: A user can require that a component with a specific property value is not in the configuration. In this case, the specified component may be absent or may be present with another value for the respective property in the solution.     
    \end{itemize}
\end{itemize}

A configuration solution is an assignment of values to component properties in the configuration problem from their domains such that the constraints are satisfied.

\begin{lstlisting}[ basicstyle=\footnotesize\ttfamily,frame = single, caption={Bike configuration problem instance as facts}, numbers=left, linewidth=\textwidth, label = {lst:facts}]
domain(bike,type,city_bike). domain(bike,type,mountain_bike).  
domain(rear_wheel,type,w1).domain(rear_wheel,type,w2).
domain(rear_wheel,type,w3).domain(front_wheel,type,w3). 
domain(front_wheel,type,w1).domain(front_wheel,type,w2).
domain(frame,type,f1).domain(frame,type,f2). domain(frame,type,f3).
domain(stand,type,s1). domain(basket,type,b1).

property_val(f1,material,aluminium).property_val(f1,basket_support,true). 
property_val(f2,material,carbon_fiber).property_val(f2,basket_support,false).
property_val(f3,basket_support,true).property_val(w1,material,aluminium).
property_val(w1,size,28).property_val(w2,size,26).property_val(w3,size,28).
property_val(w2,material,aluminium).property_val(w3,material,carbon_fiber).

mandatory_property(frame,material).

partof(bike,frame,mandatory).partof(bike,front_wheel,mandatory).
partof(bike,rear_wheel,mandatory).partof(bike,basket,optional).
partof(bike,stand,optional).

incompatible_com_pv(basket,(bike,type,mountain_bike)).
incompatible_pv_pv((front_wheel,size,26),(rear_wheel,size,28)).
incompatible_pv_pv((front_wheel,size,28),(rear_wheel,size,26)).

require_com_com(basket,stand).
require_com_pv(basket,(frame,basket_support,true)).
require_pv_pv((frame,material,aluminium),(front_wheel,material,aluminium)).

user_com(req,bike).user_com(req,basket).user_com_pv(req,(front_wheel,size,26)).
\end{lstlisting}

\section{Representing Configuration in ASP}

\subsection{Fact Format}
\begin{figure}
    \includegraphics[width=\linewidth, height=0.45\textheight]{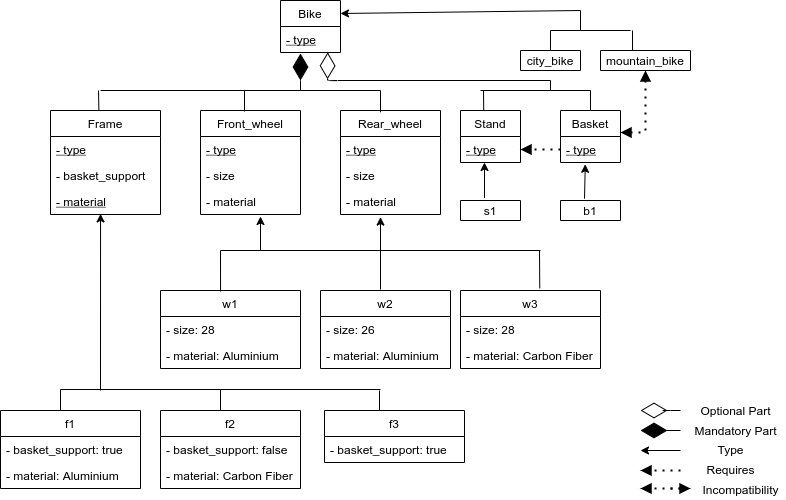}
    \caption{Bike Configuration in UML}
    \label{fig:bike}
\end{figure}

Listing \ref{lst:facts} shows an instance of the bike configuration problem shown in Figure \ref{fig:bike}.
The predicate \texttt{domain(C,P,V)} expresses that some value \texttt{V} can be assigned to an property \texttt{P} of a component \texttt{C}.
Initially, domains of the type property of all components are provided in Lines 1-6.
The predicate \texttt{property\_val(T,P,V)} shown in Line 8-12 denotes that \texttt{V} is the predefined value of the property \texttt{P} of type \texttt{T}. 
This predicate is also used to generate the domains of component properties apart from type during preprocessing.

The predicate \texttt{mandatory\_property(C, P)} in Line 14 is used to express the mandatory property constraint \textbf{A4}, where \texttt{P} is a mandatory property of a component \texttt{C}. 
During the preprocessing section of the encoding, the property type is made a mandatory property for all components.
Partonomy of the bike is listed in Lines 16-18, where the predicate \texttt{partof(C1,C2,V)} expresses that a component \texttt{C2} is a mandatory or optional part of the component \texttt{C1} depending on the respective value of \texttt{V}.
The first fact of Line 16 expresses that the frame is a mandatory part of the bike while the fact Line 18 expresses that basket is an optional part of the bike.

Product specific incompatibility relations \textbf{I1-I3} are represented in fact format in Line 20-22.
The predicate \texttt{incompatible\_com\_pv(C1,(C2,P2,V2))}, denotes an incompatibility relation \textbf{I2} between a component \texttt{C1}) and a component property value \texttt{C2,P2,V2}.
For example, the fact in Line 20 expresses that a mountain bike can't have a basket. 
Incompatibility \textbf{I3} between component attribute values \texttt{(C1,P1,V1)} and \texttt{(C2,P2,V2)} in form of predicate \texttt{incompatible\_pv\_pv((C1,P1,V1), (C2,P2,V2))} is shown in Line 21-22, which specifies that the front wheel and rear wheel can't be of different sizes.  
It can be noted that \texttt{incompatible\_com\_com(C1,C2)} to express incompatibility between component \texttt{C1} and \texttt{C2} is not given in this use case.

The requirements between bike components \texttt{R1-R3} are given in Line 24-26. 
The condition \textbf{R1} where component \texttt{C1} requires \texttt{C2} is expressed by predicate \texttt{require\_com\_com(C1, C2)}. 
For example, the fact in Line 24 expresses that a basket requires a stand.
Similarly, \texttt{require\_com\_pv(C1, (C2,P2,V2))} expresses \textbf{R2} where a component \texttt{C1} requires another component \texttt{C2} with property value \texttt{P2,V2}.
This example is shown in Line 25 expresses where a basket requires the frame to support a basket.
Requirement \textbf{R3} of component attribute value \texttt{(C2,P2,V2)} by \texttt{(C1,P1,V1)} is in form of predicate \texttt{require\_pv\_pv((C1,P1,V1), (C2,P2,V2))}.
For example, in Line 26, the front wheel has to be made up of aluminum if the frame is.  
Additionally, there is a predicate of form \texttt{require\_com\_pv((C1,P1,V1), C2)}, where presence of the component \texttt{C1} with property value \texttt{P1,V1} requires the component \texttt{C2}, which hasn't been used here.

Finally in Line 28, the user requirements are listed.
Here \texttt{user\_com(req, C)} means that user requires the component \texttt{C}.
In the example, the user requires a bike and a basket.
Users can also specify if they want the property \texttt{P} of component \texttt{C} to have a value \texttt{V}, in form of the predicate \texttt{user\_com(req, (C,P,V))}.
For example, the user requires the size of the front wheel should be 26 is given in the last fact of Line 28.
The above two predicates express \textbf{U1} and \textbf{U2} respectively. 
\textbf{U3} and \textbf{U4} can be expressed in a similar way by replacing \texttt{req} to \texttt{nreq} as the first arguement.

\begin{lstlisting}[ basicstyle=\footnotesize\ttfamily,frame = single,caption={Configuration Problem Encoding}, linewidth=\textwidth, label = {lst:encoding}, numbers=left]
domain(C,P,V) :- domain(C,type,T), property_val(T,P,V).
mandatory_property(C,type) :- domain(C,_,_).
require_com_com(Part,Whole) :- partof(Whole,Part,_).
require_com_com(Whole,Part) :- partof(Whole,Part,mandatory).

{assign(C,P,V): domain(C,P,V)}.
component(C) :- assign(C,P,V).

:- assign(C,P,V1), assign(C,P,V2), V1 < V2.

:- component(C), mandatory_property(C,P), not assign(C,P,_).

:- assign(C,type,T), assign(C,P,V), P != type , not property_val(T,P,V).
:- assign(C,type,T), property_val(T,P,V), not assign(C,P,V).

:- require_com_com(C1,C2), component(C1), not component(C2).
:- require_com_pv(C1,(C2,P,V)), component(C1), not assign(C2,P,V).
:- require_pv_com((C1,P,V),C2), assign(C1,P,V), not component(C2).
:- require_pv_pv((C1,P1,V1),(C2,P2,V2)), assign(C1,P1,V1), not assign(C2,P2,V2).

:- incompatible_com_com(C1,C2), component(C1), component(C2).
:- incompatible_com_pv(C1,(C2,P,V)), component(C1), assign(C2,P,V).
:- incompatible_pv_pv((C1,P1,V1),(C2,P2,V2)), assign(C1,P1,V1), assign(C2,P2,V2).

:- user_com(req,C), not component(C).
:- user_com_pv(req,(C,P,V)), not assign(C,P,V).
:- user_com(nreq,C), component(C).
:- user_com_pv(nreq,(C,P,V)), assign(C,P,V).
\end{lstlisting}

\subsection{Encoding and Solution}

The encoding in Listing \ref{lst:encoding} is domain-independent and is separated into three parts: preprocessing in Lines 1-4, generation in Line 6-7, and testing in the remaining part. 

In Line 1, the domains of the component properties are generated from the predefined property values of their types.
Assignment of values property of the components from their domain is given by Line 6, in form of a choice rule.
Also, to keep track of the components that are present in the configuration solution, the predicate \texttt{component(C)} is used in Line 7, where \texttt{C} is the component. 

The constraint \textbf{A4}, that all mandatory properties should be assigned a value if the component is present, is represented in Line 11. 
Along with this, the constraint \textbf{A1} is expressed using Line 2, where the type of a component is set as a mandatory property.

Constraint \textbf{A2} is represented in Line 9, meaning that a component property cannot have more than one value.
Line 13 and 14 depict rule \textbf{A3}. 
Line 13 expresses that a component cannot have property values that the assigned type doesn't have.
In addition, Line 14 expresses that the component must have all the pre-defined property values of its assigned type. 

The partonomy constraints \textbf{P1} and \textbf{P2} are mapped into requirement constraint \textbf{R1}, expressed in Line 3 and 4.
Line 3 expresses that the part of a whole component requires the whole component. 
In addition, whole components require all their mandatory components to be present is enforced in Line 4.
The component requirements are defined in Line 16-19. 
Line 16 states that a component can require another component (\textbf{R1}).
Rule \textbf{R2} is represented by Lines 17 and 18 where a component can require a certain property value from another component and vice versa. 
The last requirement case, that a component with a specific property value requires another component with a specific property value is represented by Line 19 (rule \textbf{R3}). 

The incompatibility constraints are enforced in Lines 21-23. 
Line 28 translates constraint \textbf{I1}, where a component is incompatible with another component.
\textbf{I2} is represented by Line 30. 
The last case that a component with a certain property value is incompatible with another component with a specific property value is done in Line 32 (\textbf{I3}).

The user requirements are represented in Line 25-28.
A user can require a certain component (\textbf{U1}) or a component with a specific property value (\textbf{U2}).
This is represented in Line 25 and 26 respectively. 
On the contrary, the user can also deny certain components or attribute values to be in the configuration (\textbf{U3, U4}) which is represented in Line 27 and 28.

Each predicate in the solution \texttt{assign(C,A,V)} is intended to express that the component \texttt{C} with attribute \texttt{A} and value \texttt{V} is present in the solution. 
Given the facts in Listing \ref{lst:facts} and encoding in Listing \ref{lst:encoding}, the solution is represented in Listing \ref{lst:solution}.
Since the user needs a basket that is incompatible with the mountain bike, the user is assigned a city bike. 
Due to requirements in Lines 24 and 25 in Listing \ref{lst:facts}, stand is added to the configuration, and f1 is selected as the frame.
The wheel w2 is used as front wheel because of user requirement.
Due to incompatibility expressed in Line 21 of Listing \ref{lst:facts}, w2 is also used as the rear wheel.

\begin{lstlisting}[ basicstyle=\footnotesize\ttfamily,frame = single, caption={Bike Configuration Solution}, linewidth=\textwidth, label = {lst:solution}, numbers=left]
assign(bike,type,city_bike). assign(rear_wheel,type,w2).
assign(front_wheel,type,w2). assign(frame,type,f1). 
assign(stand,type,s1). assign(front_wheel,material,aluminium).  
assign(frame,basket_support,true). assign(basket,type,b1).
assign(frame,material,aluminium). assign(front_wheel,size,26). 
assign(rear_wheel,size,26). assign(rear_wheel,material,aluminium)

\end{lstlisting}

 \section{Future Work}\label{sec:future}

This is a preliminary work with the final objective of developing an interactive product configuration system \cite{fahakrscscta20a} \cite{hedajade17a}. 
Identification and encoding of more concepts such as resource  constraints, default values, user preferences among others would be the next step.
Mapping from commonly used graphical and constraint based configuration representation methods to ASP should also be done.
Interactive configuration has many construction zones \cite{zhang14a}, including product recommendation, configuration diagnosis \cite{junker04a} and explanation. 
Further investigation of these techniques in the context of ASP is another objective.
Finally, a major step would be implementation of these interactive techniques using clingo \cite{gekakasc14b} along with its sophisticated Python API that allows a fine-grained handling of the solving process.

\bibliographystyle{eptcs}

\end{document}